\documentclass[10pt,twocolumn,letterpaper]{article}

\usepackage{iccv}
\usepackage{times}
\usepackage{epsfig}
\usepackage{graphicx}
\usepackage{amsmath}
\usepackage{multirow}
\usepackage{amssymb}
\usepackage{enumitem}

\usepackage[pagebackref=true,breaklinks=true,letterpaper=true,colorlinks,bookmarks=false]{hyperref}

\iccvfinalcopy 

\def\iccvPaperID{4925} 

\ificcvfinal\fi

\begin{document}

\title{A Benchmark for Chinese-English Scene Text Image Super-resolution}

\author{Jianqi Ma\textsuperscript{1,2}, Zhetong Liang\textsuperscript{2}, Wangmeng Xiang\textsuperscript{1}, Xi Yang\textsuperscript{1,2}, Lei Zhang\textsuperscript{1,2}\\
\textsuperscript{1}The Hong Kong Polytechnic University; \textsuperscript{2}OPPO Research\\
{\tt\footnotesize \{csjma, cswmxiang, cslzhang, csxyang\}@comp.polyu.edu.hk,}
{\tt\footnotesize zhetongliang@163.com}
}

\maketitle
\ificcvfinal\thispagestyle{empty}\fi

\begin{abstract}
   Scene Text Image Super-resolution (STISR) aims to recover high-resolution (HR) scene text images with visually pleasant and readable text content from the given low-resolution (LR) input. Most existing works focus on recovering English texts, which have relatively simple character structures, while little work has been done on the more challenging Chinese texts with diverse and complex character structures. In this paper, we propose a real-world Chinese-English benchmark dataset, namely Real-CE, for the task of STISR with the emphasis on restoring structurally complex Chinese characters. The benchmark provides 1,935/783 real-world LR-HR text image pairs~(contains 33,789 text lines in total) for training/testing in 2$\times$ and 4$\times$ zooming modes, complemented by detailed annotations, including detection boxes and text transcripts. Moreover, we design an edge-aware learning method, which provides structural supervision in image and feature domains, to effectively reconstruct the dense structures of Chinese characters. We conduct experiments on the proposed Real-CE benchmark and evaluate the existing STISR models with and without our edge-aware loss. The benchmark, including data and source code, is available at \href{https://github.com/mjq11302010044/Real-CE}{https://github.com/mjq11302010044/Real-CE}.
\end{abstract}

\section{Introduction}

\label{sec:intro}

Text images are different from natural images in that the main contents are composed of words and characters to express different meanings and ideas. Due to limited sensor resolution and long photographing distance, the captured text images often have degraded quality with blurry and noisy contents, impairing the readability of the text. Therefore, scene text image super-resolution (STISR) is demanded to reconstruct clear and legible text contents.

STISR has long been studied in the computer vision community~\cite{peyrard2015icdar2015,xu2017learning,wang2019textsr,quan2020collaborative}. The traditional STISR methods investigate various priors on text restoration and hand-craft the text super-resolution process~\cite{brown2004image,chen2011effective}. Since the manually designed priors cannot represent the complex text structures and degradation process, the traditional methods have limited performance. Deep learning based STISR methods train convolutional neural networks (CNNs) on datasets with low-resolution (LR) and high-resolution (HR) text image pairs, which can learn the complex text priors through data and reconstruct high-quality text images.


In deep learning based STISR~\cite{peyrard2015icdar2015,xu2017learning,wang2019textsr,quan2020collaborative}, datasets play an important role in model training and evaluation, because the image pairs encode the text transformation from low to high resolution. In the early stage, synthetic datasets are widely used~\cite{peyrard2015icdar2015,xu2017learning,quan2020collaborative}, in which high-quality text images are collected as HR ground truths, and the LR images are generated by imposing synthetic degradations (\eg, bicubic downsampling or blurring) on the HR images. Since the real-world degradations are quite different from the synthetic ones, the STISR models trained on the synthetic datasets have limited performance on real-world LR text images. To alleviate this problem, Wang~\etal~\cite{wang2020scene} built a real-world text image dataset called TextZoom. The LR and HR text images in TextZoom are captured with different camera focal lengths and undergo the real-world degradation process. TextZoom provides a benchmark for the STISR task, which allows standardized evaluation of STISR methods in terms of text recognition precision.


\begin{figure*}[t]
  \centering
  \includegraphics[width=0.85\linewidth]{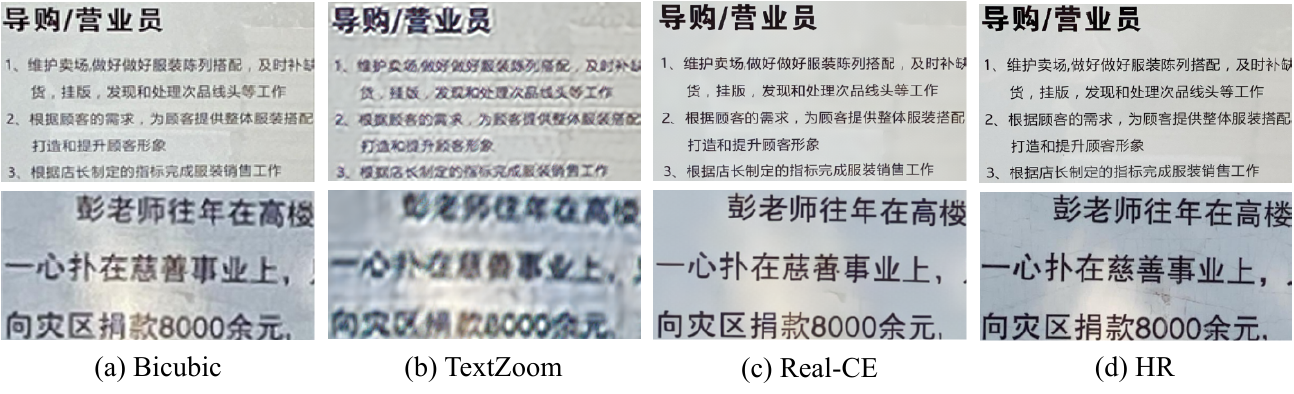}
  \caption{Comparison of STISR results on Chinese text images by methods trained on TextZoom and Real-CE datasets. From left to right are (a)~bicubic LR images, STISR outputs by RRDB model~\cite{wang2018esrgan} trained on (b) TextZoom~\cite{wang2020scene} and (c) our Real-CE, and (d) the ground-truth HR text images. Please zoom in for more details.}
  \label{fig:tissue}
\end{figure*}

Though the TextZoom dataset has largely facilitated the research of real-world STISR~\cite{wang2020scene,chen2021scene,ma2021text,chen2021text,zhao2021scene,ma2022text,zhao2022c3}, it has some limitations. First, TextZoom only contains English texts composed of limited number of characters (\ie, $26$ letters) with simple stroke structures. As a result, models trained on TextZoom will produce inferior results on structurally complex characters like Chinese. Examples are shown in Figure~\ref{fig:tissue}(b). One can see that the model trained on TextZoom produces visually unpleasant artifacts on the reconstructed Chinese texts. This is because Chinese texts have a much larger number of characters, and many of them have complex structures. Thus, it is a more challenging task for performing STISR on Chinese texts. Moreover, TextZoom focuses on small and fixed-size text images~(\ie, $32\times128$), and thus the models trained on TextZoom cannot generalize to texts with various resolutions. Therefore, new dataset and benchmark are highly demanded for the research of STISR on Chinese text images.

To tackle the above-mentioned problems, in this work we develop a novel real-world Chinese-English benchmark dataset, termed Real-CE, for the training and evaluation of STISR models on both Chinese and English texts. The benchmark provides $1,935$ real-world LR-HR image pairs for training, and $783$ for testing ($261$ and $522$ pairs for $4\times$ and $2\times$ zooming modes, respectively). It contains $24,666$ Chinese text lines and $9,123$ English text lines in total with different sizes. Detailed annotations on the image pairs, including detection boxes and text transcripts, are also provided to assist the training and evaluation. We also design the evaluation process to adapt to different sizes of text lines, aiming to preserve the visual quality of SR text images from resizing. Furthermore, we propose an edge-aware learning method for the reconstruction of Chinese texts with complex stroke structures. The text edge map is introduced as the network input as well as a structural loss in the training process, enhancing the learning on text structural regions. Experimental results show that models trained on our Real-CE data achieve superior performance over TextZoom on Chinese text super-resolution (as shown in Figure~\ref{fig:tissue}(c)) and the edge-aware learning can further promote the reconstruction quality on text regions. 

The paper is organized as follows. Section 2 reviews the works on STISR research. Section 3 introduces the Real-CE benchmark in detail. Section 4 describes the edge-aware learning method. Section 5 shows the experimental results on the benchmark and Section 6 concludes the paper.

\section{Related Work}
Our work is related to single image super-resolution (SISR), scene text image super-resolution (STISR), and English and Chinese text recognition, as reviewed below.

\textbf{SISR.} SISR estimates a high-resolution (HR) output by intaking the low-resolution (LR) image as input. Traditional approaches apply manually-designed priors for this task in terms of statistical information~\cite{GunturkAM04}, self-similarity~\cite{MairalBPSZ09} and sparsity~\cite{YangWHM10}. Recent deep-learning methods employ convolutional neural networks (CNNs) for SISR and achieve significantly better performance. As a pioneer work, SRCNN~\cite{dong2015image} adopts a three-layer CNN to perform HR estimation. Later on, more elaborate designs on network architecture further upgrade the SISR performance, including residual connection~\cite{lim2017enhanced}, Laplacian pyramid~\cite{lai2017deep}, dense connection block~\cite{wang2018esrgan} and the Transformer architecture~\cite{liang2021swinir,zhang2022efficient}. Adversarial learning techniques have also been applied for more photo-realistic results~\cite{ledig2017photo,wang2018recovering}.


\textbf{STISR benchmarks and methods.} STISR focuses on scene text images. It aims to reconstruct the text shape by upgrading the image resolution in order to benefit the down-stream recognition task. The early methods of STISR directly adopt the CNN architectures used in general SISR tasks. In~\cite{dong2015boosting}, Dong \etal~adopted SRCNN~\cite{dong2015image} to text images, and achieved state-of-the-art performance in ICDAR 2015 competition~\cite{peyrard2015icdar2015}. PlugNet~\cite{mouplugnet} employs a pluggable super-resolution unit to learn the semantics in LR images in feature domain. TextSR~\cite{wang2019textsr} utilizes the text recognition loss to supervise SR recovery learning and improve the text recognition. 
Aiming to learn text image deblurring and super-resolution, Xu~\etal~\cite{xu2017learning} and Quan~\etal~\cite{quan2020collaborative} collected high-quality document text data to evaluate synthetic image deblurring and super-resolution.

To address real-world STISR problems, Wang~\etal \cite{wang2020scene}~built an STISR benchmark, namely TextZoom, which provides the LR and HR text image pairs extracted from real-world SISR datasets~\cite{zhang2019zoom,cai2019toward}. They also proposed TSRN~\cite{wang2020scene} by applying the sequential residual block to model the sequential semantics in image features. SCGAN~\cite{xu2017learning} adopts GAN loss to supervise the STISR model for more realistic text images. Quan \etal~\cite{quan2020collaborative} proposed a cascading network for reconstructing high-quality text images in both high-frequency domain and image domain. Chen \etal~\cite{chen2021scene,chen2021text}, Zhao~\etal~\cite{zhao2021scene} and Ma~\etal~\cite{ma2022text} upgraded the network block structures to enhance the STISR performance with transformer-based networks or text prior. 

However, current STISR methods are designed only for English-based text line images~(\ie, the TextZoom) with many limitations. We therefore make an attempt to build a bilingual benchmark to fill in the blank.


\textbf{Scene text recognition.} Scene text recognition~(STR) aims to recognize the semantic meaning in the text image by predicting the characters or the whole word~\cite{jaderberg2014deep,he2015reading,jaderberg2016reading,liu2016star}. It can be considered as an image-to-sequence problem. CRNN~\cite{shi2016end} uses recurrent neural networks to model semantic information. Recently, attention-based methods have achieved great success due to their robustness against shape variations of text images~\cite{cheng2017focusing,cheng2018aon,shi2018aster}. However, most methods are proposed for English text, and Chinese scene text recognition receives less attention~\cite{du2016deep,hu2017end,chen2017compact}. To promote research along this line, Chen~\etal~\cite{chen2021benchmarking} attempted to benchmark the Chinese scene text recognition with unified input and evaluation metrics.
In this paper, text recognizers in both languages are adopted for evaluating text recognition after STISR.

\section{Real-CE Benchmark}

The proposed benchmark includes a dataset with Chinese and English LR-HR text image pairs and an evaluation protocol with five metrics.



\subsection{Dataset Construction}
\begin{figure}[t]
  \centering
  \includegraphics[width=\linewidth]{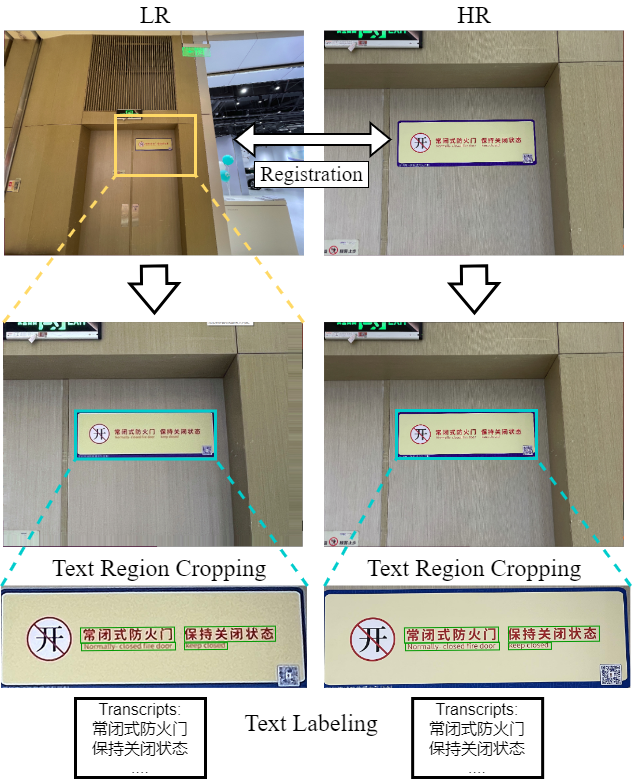}
  \caption{The pipeline of data processing. From top to bottom: the center area of the LR image is first registered to the HR image, then the corresponding text regions in the HR and LR images are cropped and manually aligned; finally, the text lines are annotated and the transcripts are labeled. }
  \label{fig:Cropping}
  \vspace{-0.3cm}
\end{figure}

The dataset is constructed by several steps, including data collection, registration, text cropping and text labeling, which are illustrated in Figure~\ref{fig:Cropping}.

\begin{figure}[t]
  \centering
  \includegraphics[width=\linewidth]{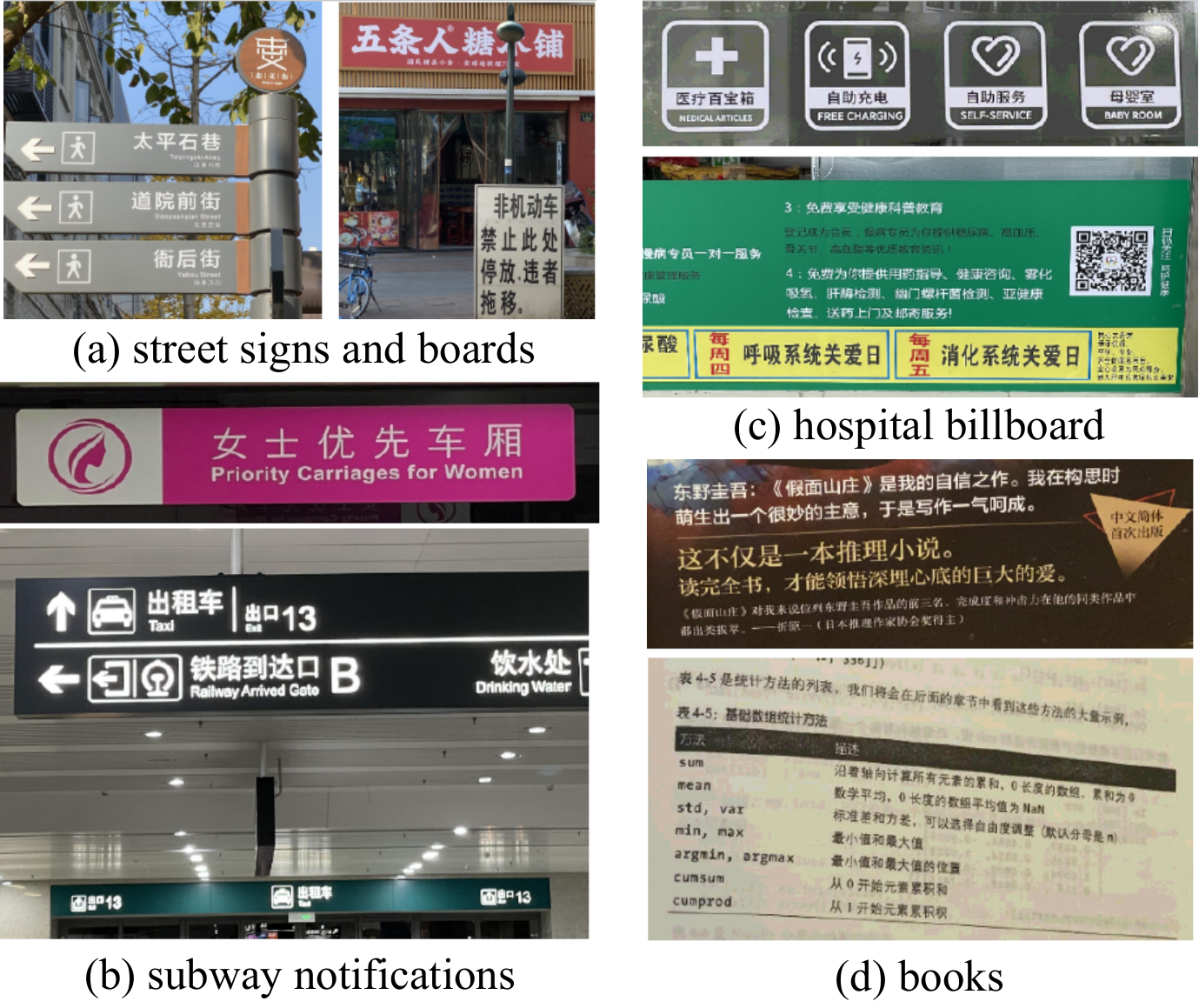}
  \caption{Typical scenes in our collected Real-CE dataset.}
  \label{fig:scenes_and_resources}
  \vspace{-0.3cm}
\end{figure}

\textbf{Data collection.} We adopt iPhone 11 pro and iPhone 12 pro for text image collection. Both of them are equipped with camera modules of three fixed focal lengths ($13$~mm, $26$~mm and $52$~mm), which allow us to capture the same scene with different focal lengths simultaneously. Image pairs collected by these devices enable the training of STISR models in $2\times$ (from $13$~mm to $26$~mm, and from $26$~mm to $52$~mm) and $4\times$ (from $13$~mm to $52$~mm) zooming modes. We capture the Chinese and English text images from various scenes and resources, including band and curve outdoor street signboards, subway notifications, deformed books and hospital billboards, so that the diversity of text contents, presentation and lighting conditions can be ensured. Since the three cameras may have different image processing pipelines, we use CameraPixels app~\footnote{https://apps.apple.com/us/app/camerapixels-pro/id1148178499} to align the colors and brightness of the three captured images. Figure~\ref{fig:scenes_and_resources} shows some typical scenes in the collected dataset.

\textbf{Image registration.} We adopt the image registration method proposed in Cai~\etal~\cite{cai2019toward} for the alignment of LR and HR text image pairs. Specifically, we take the images captured by $52$mm lens as the ground-truth HR images since they have the best quality, and iteratively register shorter-lens LR counterparts to it. The algorithm also enables finer-grained adjustment, which reduces the color and brightness differences between the LR-HR image pairs.

\textbf{Text region cropping.}
Though the text images are centered on text content, they still contain a large proportion of background area. We therefore crop the central text region from the LR and HR images to exclude background areas, followed by a manual adjustment to ensure accurate LR-HR image pair alignment. 



\textbf{Text labeling.}
Besides the HR ground truths, we provide two extra text labels, including detection boxes and text transcripts. The detection boxes provide the location of the text areas, while the text transcripts record the semantics of the texts. For the detection box annotation, we first apply some text detection methods~(\eg, RRPN~\cite{ma2018arbitrary,ma2020rrpn}) to provide a coarse detection result, then we refine the detection results manually to provide a precise boundary of each text line in the cropped text region. For the text transcripts annotation, we employ a text recognizer pretrained in Chinese and English~\cite{shi2016end,chen2021benchmarking} to obtain the initial transcripts, followed by a manual refinement. With precise text labeling, STISR models can be evaluated from the aspect of text recognition on the test set of our Real-CE benchmark.


\subsection{Dataset Statistics}

Our dataset contains both $2\times$ and $4\times$ zooming modes for training and testing. The detailed statistics of our dataset are shown in Table~\ref{table:Splitting}.  Our Real-CE dataset contains $33,789$ text line pairs. In particular, $24,666$ of them are Chinese texts while the rest are English texts. 

{\bf Text region pairs.} Our dataset contains $2,718$ text region pairs, $1,935$ of which are training pairs and the rest are testing pairs. Among the testing pairs, there are $261$ pairs for $4 \times$~($13$mm to $52$mm~) zooming and another $522$ pairs for $2 \times$~($26$mm to $52$mm and $13$mm to $26$mm~) zooming evaluation. All the cropped HR text regions are ranged from size of $228 \times 396$ to $4,032 \times 3,024$. Each text region contains one or more text lines. 

{\bf Text lines.} The text semantics and language are distinguished with text lines. Text boxes and recognition are also annotated by lines. There are $23,547$ text lines for training, $3,414$ text lines for $4\times$ zooming evaluation, and $6,828$ text lines for $2\times$ zooming evaluation. The size of text lines ranges from $16 \times 22$ to $1,156 \times 2,883$. The category of the characters in Real-CE is $3,755$ in total.
\begin{table}
\small
\centering
\begin{tabular}{|l|p{1.0cm}|p{1.0cm}|p{1.0cm}|p{1.0cm}|}
\hline
     & \multicolumn{2}{c|}{Region} & \multicolumn{2}{c|}{Text line} \\\hline
 SR factor & \multicolumn{1}{c|}{$4\times$} & \multicolumn{1}{c|}{$2\times$} & \multicolumn{1}{c|}{$4\times$} & \multicolumn{1}{c|}{$2\times$}\\\hline
 train & \multicolumn{1}{c|}{~$645$~~} & \multicolumn{1}{c|}{$1,290$} & \multicolumn{1}{c|}{$7,849$} & \multicolumn{1}{c|}{$15,698$}\\\hline
 test & \multicolumn{1}{c|}{~$261$~~} & \multicolumn{1}{c|}{$522$} & \multicolumn{1}{c|}{$3,414$} & \multicolumn{1}{c|}{$6,828$}\\\hline
 Max resolution & \multicolumn{2}{c|}{$4,032 \times 3,024$} & \multicolumn{2}{c|}{$1,156 \times 2,883$} \\\hline
 Min resolution & \multicolumn{2}{c|}{$228 \times 396$} & \multicolumn{2}{c|}{$16 \times 22$}\\\hline
 Chinese & \multicolumn{2}{c|}{-} & \multicolumn{1}{c|}{$8,222$} & \multicolumn{1}{c|}{$16,444$} \\\hline
 English & \multicolumn{2}{c|}{-} & \multicolumn{1}{c|}{$3,041$} & \multicolumn{1}{c|}{$6,082$} \\\hline
\end{tabular}
\vspace{0.3cm}
\caption{Statistics of the constructed Real-CE dataset.}
\label{table:Splitting}
\end{table}%

\subsection{Evaluation Protocol}
\label{sec:eva}

\begin{figure}[t]
  \centering
  \includegraphics[width=0.85\linewidth]{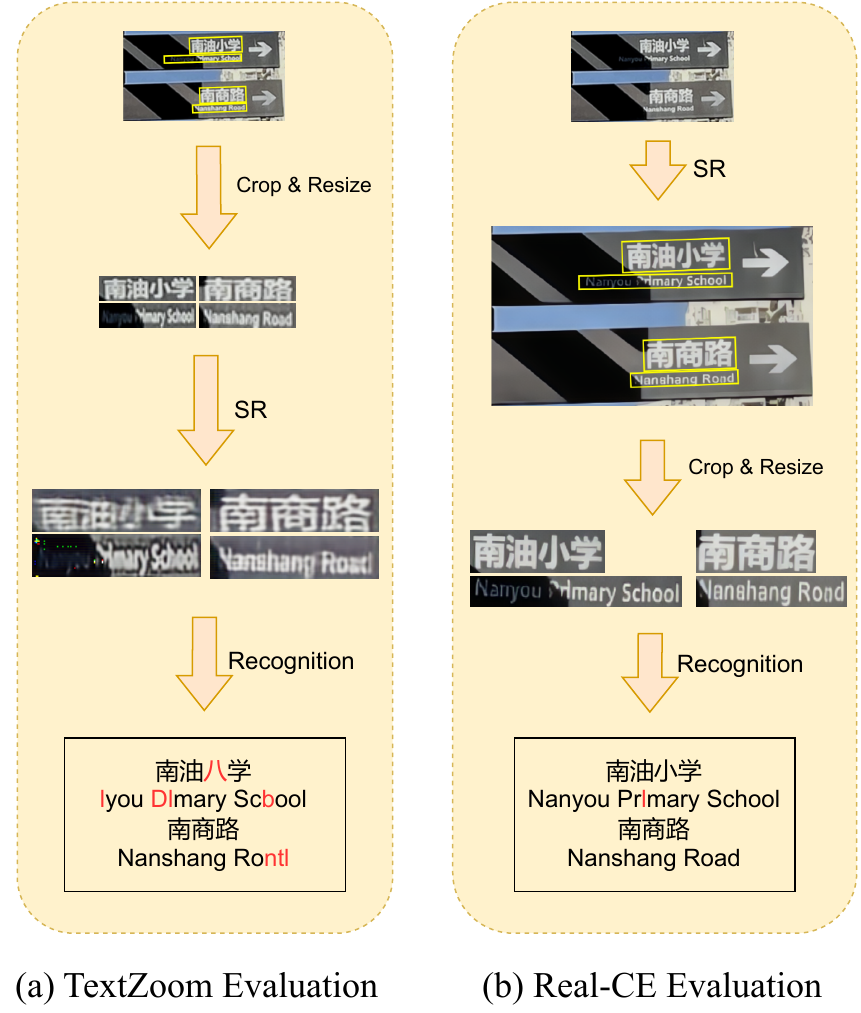}
  \caption{Comparison of the TextZoom evaluation and Real-CE evaluation protocols. Wrong recognition results are in red. Please zoom in for more details.}
  \label{fig:teaser_benchmark}
  \vspace{-0.3cm}
\end{figure}

To evaluate the performance of STISR models on Real-CE, we employ $5$ metrics, including structural similarity index measure~(SSIM)~\cite{wang2004image}, peak signal-to-noise ratio~(PSNR), learned perceptual image patch similarity (LPIPS)~\cite{zhang2018unreasonable}, normalized edit distance~(NED) and word accuracy~(ACC). Among them, PSNR, SSIM and LPIPS measure the errors between the reconstructed HR images and the ground truths. In particular, PSNR and SSIM are evaluated in image space while LPIPS is evaluated in feature space. ACC and NED employ text recognition models to evaluate the recognition accuracy of the reconstructed HR images. Here we adopt pre-trained CRNN~\cite{shi2016end,chen2021benchmarking} as the text recognition model for evaluation. Particularly, ACC computes the word-level accuracy of the predicted sequence. NED between the predicted text sequence $P$ and the ground truth text image label $G$ are computed as follows:
\begin{equation}
\begin{array}{l}
NED(P, G) = 
\left.
            1 - \cfrac{ED(P, G)}{max(|P|, |G|)}
\right.
\end{array}
\label{equ_NED},
\end{equation}
where $ED(\cdot)$ stands for the edit distance calculation, $|P|$ and $|G|$ refer to the length of the prediction and the ground-truth label. Therefore, the predicted sequence is more accurate and closer to the ground-truth label when the NED is larger. When we measure long texts, the ACC index may not fully reflect the recognition correctness at character level, while NED can measure it in a finer-grained manner.

In the testing process, the trained STISR models are performed on the original LR text region image to obtain the reconstructed HR images first. Then the text lines in reconstructed HR images are cropped and kept in their original ratio for recognition evaluation in terms of ACC and NED. The evaluation process is illustrated in Figure~\ref{fig:teaser_benchmark}~(b). 
Compared with the evaluation protocol of TextZoom~\cite{wang2020scene} (see Figure~\ref{fig:teaser_benchmark}~(a)), which trains and evaluates text lines with fixed sizes and shapes, our protocol can avoid the text deformation brought by the resizing operation. As shown in Figure~\ref{fig:teaser_benchmark}~(a), such an arrangement is unfriendly for Chinese long text in Real-CE~(often presented as sentences), resulting in low reconstruction quality and recognition accuracy.


\section{Text Edge-aware STISR}


Different from English characters, Chinese characters are composed of more basic radical-level parts~(one can refer to~\cite{chen2021benchmarking} for more details) and have more complicated internal structures. Therefore, elaborated designs are needed to enhance the model capacity for Chinese text reconstruction. In this section, we propose an edge-aware learning method, which uses the text edge map as input and an edge-aware loss for supervision.

\subsection{Text Edge Map} 
The text information in an image is inevitably blended with complex background. This will weaken the saliency of the text structures and somehow impairs the text reconstruction process. Text edge information is helpful to tackle this problem because it can effectively guide an STISR model to be better aware of the text structures and strokes.

\begin{figure}[t]
  \centering
  \includegraphics[width=\linewidth]{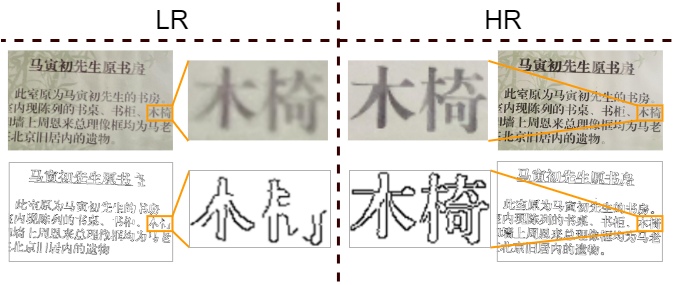}
  \caption{LR-HR RGB images (top) and their Canny edge maps (bottom). Foreground edges are drawn in black, and background in white for better visualization.}
  \label{fig:CannyEdge}
\end{figure}

\begin{figure}[t]
  \centering
  \includegraphics[width=\linewidth]{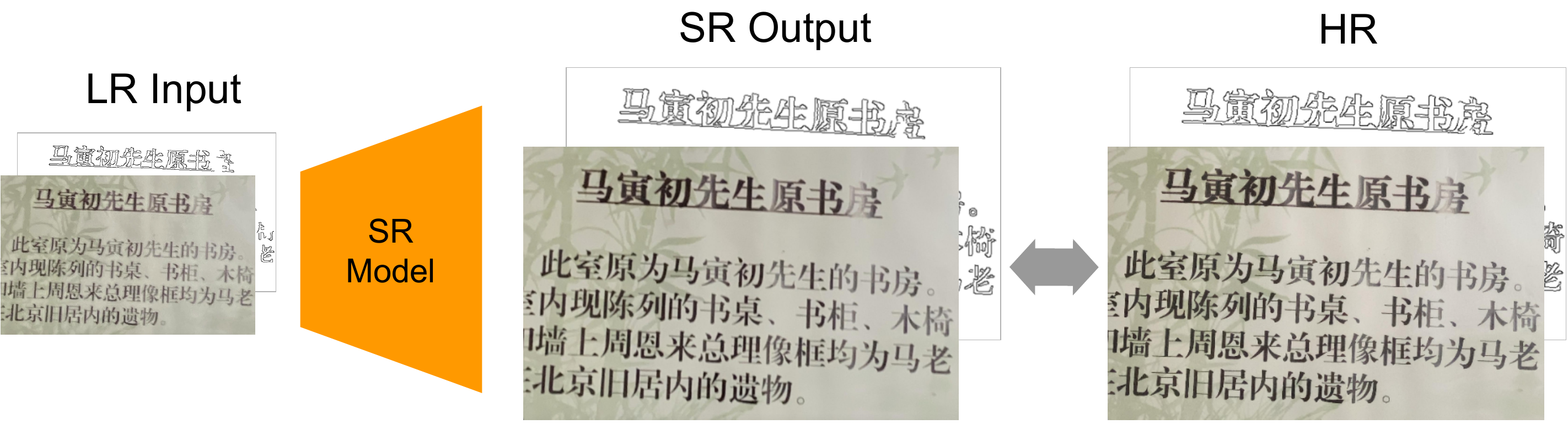}
  \caption{Illustration of the edge-aware STISR model learning. The edge map of the LR image is extracted and input to the network, and the edge map of the HR image is used to supervise the network training.}
  \label{fig:CannyLearning}
  \vspace{-0.3cm}
\end{figure}

We adopt the Canny edge detector~\cite{canny1986computational} to compute a text edge map, denoted as $\mathcal{C}$, in the training process. The text edge map assigns value $1$ to the text contour area and $0$ to the background. Thus, the text edge map contains text structures and excludes the background information. From Fig.~\ref{fig:CannyEdge}, one can see that the character shape and structure may be unclear in the LR-HR image pairs, while in their Canny edge maps, the text shapes and structures are enhanced. We compute edge maps for both LR and HR images in the dataset. The LR edge map $\mathcal{C}_{LR}$ is concatenated with the LR image in channel dimension as the network input, which is shown in Fig.\ \ref{fig:CannyLearning}. With this extra input, the STISR model can learn a stronger feature representation of the finer-grained text structure.

\subsection{Edge-aware Loss} 

We propose an edge-aware loss based on the computed edge map. First, the STISR model is modified to output both the reconstructed HR text image $\hat{\mathcal{I}}_H$ and an estimated HR text edge map $\hat{\mathcal{C}}_H$. This estimated text edge map is used in the training stage to gain extra supervision, but is discarded in the testing stage. The EA loss is computed between the estimated text edge map and the ground truth edge map at pixel level and feature level.

At pixel level, we adopt the $\mathcal{L}_1$ loss in image domain between the estimated HR edge map $\hat{\mathcal{C}}_{H}$ and the ground truth HR text map $\mathcal{C}_{H}$. Therefore, the EA loss at the pixel level $\mathcal{L}^{P}_{EA}$ is calculated as:
\begin{equation}
    \mathcal{L}^{P}_{EA} = |\mathcal{C}_{H} - \hat{\mathcal{C}}_{H}|.
\end{equation}

\begin{table*}[t]
\footnotesize
\centering
\setlength{\tabcolsep}{5pt}
\scalebox{1.0}[0.9]{
\begin{tabular}{c|l|l|ccccc|ccccc}
\hline\hline
& & SR factor & \multicolumn{5}{c|}{$4\times$} & \multicolumn{5}{c}{$2\times$}\\\hline
\multicolumn{2}{c|}{Approach} & train set & PSNR $\uparrow$ & SSIM $\uparrow$ & LPIPS $\downarrow$ & ACC $\uparrow$ & NED $\uparrow$ & PSNR $\uparrow$ & SSIM $\uparrow$ & LPIPS $\downarrow$ & ACC $\uparrow$ & NED $\uparrow$\\\hline
& \footnotesize{Bicubic} &  & 19.65 & 0.6684 & 0.3987 & 0.2759 & 0.6173 & 20.82 & 0.7106 & 0.2100 & 0.3475 & 0.6982\\\cline{1-13}
\multirow{15}{*}{\rotatebox{90}{SISR Methods}} & \multirow{3}{*}{\footnotesize{SRRes~\cite{ledig2017photo}}} & \footnotesize{TZ~\cite{wang2020scene}} & 19.72 & 0.6808 & 0.3872 & 0.2201 & 0.5992 & 20.28 & 0.6762 & 0.3467 & 0.2742 & 0.6401\\
& & \footnotesize{RS~\cite{cai2019toward}}  &  18.60 & 0.6576 & 0.3736 & 0.2642 & 0.6087 &  19.10 & 0.6872 & 0.3244 & 0.2977 & 0.6671\\
& & \footnotesize{RC}  & \textbf{20.22} & \textbf{0.7224}  & \textbf{0.2665} & \textbf{0.2879} & \textbf{0.6361} & \textbf{20.72} & \textbf{0.7360}  & \textbf{0.2116} & \textbf{0.3499} & \textbf{0.6996}\\\cline{2-13}
& \multirow{3}{*}{\footnotesize{RRDB~\cite{wang2018esrgan}}} & \footnotesize{TZ~\cite{wang2020scene}} & 18.95 & 0.6575 & 0.4495 & 0.1463 & 0.3776 & 19.43 & 0.6899 & 0.3887 & 0.1962 & 0.4665\\
&  & \footnotesize{RS~\cite{cai2019toward}} & 19.59 & 0.6703 & 0.2765 & 0.2590 & 0.6201 & 20.34 & 0.7312 & 0.2267 & 0.3307 & 0.6772\\
&  & \footnotesize{RC}  & \textbf{20.23} & 0.\textbf{7231}  & \textbf{0.2626} & \textbf{0.2920} & \textbf{0.6421} & \textbf{21.10} & \textbf{0.7535}  & \textbf{0.2065} & \textbf{0.3494} & \textbf{0.7003}\\\cline{2-13}
& \multirow{3}{*}{\footnotesize{EDSR~\cite{lim2017enhanced}}} & \footnotesize{TZ~\cite{wang2020scene}}  & 18.88 & 0.6512  & 0.4860 & 0.0799 & 0.3414 & 19.32 & 0.6904  & 0.4012 & 0.1369 & 0.4077 \\
& & \footnotesize{RS~\cite{cai2019toward}}  & 19.59 & 0.6728  & 0.3295 & 0.2702 & 0.6231 & 20.02 & 0.7291 & 0.2837 & 0.3189 & 0.6708\\
&  & \footnotesize{RC} & \textbf{20.16} & \textbf{0.7195}  & \textbf{0.2883} & \textbf{0.2882} & \textbf{0.6330} & \textbf{20.74} & \textbf{0.7448}  & \textbf{0.2258} & \textbf{0.3468} & \textbf{0.6954} \\\cline{2-13}
 & \multirow{3}{*}{\footnotesize{RCAN~\cite{zhang2018image}}} & \footnotesize{TZ~\cite{wang2020scene}} & 18.97 & 0.6277  & 0.4816 & 0.0810 & 0.3424 & 19.48 & 0.6488  & 0.4075 & 0.1507 & 0.5420  \\
& & \footnotesize{RS~\cite{cai2019toward}} & 19.55 & 0.6661  & 0.3475 & 0.2450 & 0.5989 & 20.05 & 0.7044 & 0.2806 & 0.2968 & 0.6776 \\
&  & \footnotesize{RC} & \textbf{20.33} & \textbf{0.7232}  & \textbf{0.2878} & \textbf{0.2879} & \textbf{0.6321} & \textbf{20.98} &\textbf{ 0.7435}  & \textbf{0.2173} & \textbf{0.3484} & \textbf{0.7006} \\\cline{2-13}
&  \multirow{3}{*}{\footnotesize{ELAN~\cite{zhang2022efficient}}} & \footnotesize{TZ~\cite{wang2020scene}} & 19.21 & 0.6459 & 0.3796 & 0.1778 & 0.4764 & 20.10 & 0.6653 & 0.3241 & 0.2254 & 0.5467 \\
& & \footnotesize{RS~\cite{cai2019toward}} & 19.60 & 0.6660 & 0.3348 & 0.2674 & 0.6228 & 20.48 & 0.6907 & 0.2732 & 0.3104 & 0.6642  \\
& & \footnotesize{RC}  & \textbf{20.39} & \textbf{0.7299} & \textbf{0.2892} & \textbf{0.2953} & \textbf{0.6404} & \textbf{21.16} & \textbf{0.7480} &\textbf{ 0.2201} & \textbf{0.3508} & \textbf{0.6992} \\\hline\hline

\multirow{12}{*}{\rotatebox{90}{STISR Methods}} & \multirow{3}{*}{TSRN~\cite{wang2020scene}} & TZ~\cite{wang2020scene} & 17.47 & 0.4853 & 0.1990  & 0.1796 & 0.3874 & 18.73 & 0.5676 & 0.1855 & 0.2471 & 0.4622\\
& & RS~\cite{cai2019toward} & 17.83 & \textbf{0.4899} & 0.2154 & 0.1733 & 0.3759 & \textbf{19.06} & \textbf{0.5322} & 0.1892 & 0.2675 & 0.4526\\
& & RC & \textbf{18.11} & 0.4850 & \textbf{0.1981} & \textbf{0.2316} & \textbf{0.4159} & 18.99 & 0.5233 & \textbf{0.1677} & \textbf{0.2854} & \textbf{0.4809} \\\cline{2-13}

& \multirow{3}{*}{TPGSR~\cite{ma2021text}} & TZ~\cite{wang2020scene} & 17.37 & \textbf{0.4913} & 0.1896 & 0.2076 & 0.3842 & 17.99 & 0.5312 & 0.1686 & 0.2655 & 0.4423 \\
& & RS~\cite{cai2019toward} & 17.65 & 0.4772 & 0.1947 & 0.2203 & 0.3930 & 18.56 & 0.5462 & 0.1754 & 0.2952 & 0.4658  \\
& & RC & \textbf{18.07} & 0.4758 & \textbf{0.1843} &\textbf{0.2326} & \textbf{0.4123} & \textbf{18.83} & \textbf{0.5562} & \textbf{0.1661} & \textbf{0.3007} & \textbf{0.4913}\\\cline{2-13}

& \multirow{3}{*}{TBSRN~\cite{chen2021scene}} & TZ~\cite{wang2020scene} & 17.59 & \textbf{0.4919} & 0.1767 & 0.2246 & 0.4133 & 18.41 & \textbf{0.5456} & \textbf{0.1588} & 0.2905 & 0.4896\\
& & RS~\cite{cai2019toward} & 17.69 & 0.4762 & 0.1849 & 0.2235 & 0.4021 & 18.69 & 0.5309 & 0.1666 & 0.2895 & 0.4644  \\
& & RC & \textbf{18.33} & 0.4826 & \textbf{0.1715} & \textbf{0.2527} & \textbf{0.4444} & \textbf{19.01} & 0.5366 & 0.1652 & \textbf{0.3181} & \textbf{0.5294}  \\\cline{2-13}

& \multirow{3}{*}{TATT~\cite{ma2022text}} & TZ~\cite{wang2020scene} & 17.43 & \textbf{0.5010} & 0.2003 & 0.2100 & 0.3926 & 18.24 & 0.5667 & 0.1827 & 0.2755 & 0.4993  \\
& & RS~\cite{cai2019toward} & 17.66 & 0.4993 & 0.2256 & 0.2092 & 0.3916 & 18.47 & 0.5253 & 0.1930 & 0.2749 & 0.4702  \\ 
& & RC & \textbf{17.96} & 0.4904 & \textbf{0.1804} & \textbf{0.2330} & \textbf{0.4342} & \textbf{19.06} & \textbf{0.5772} & \textbf{0.1590} & \textbf{0.3127} & \textbf{0.5240}  \\\hline
 

&  HR & & - & - &  - & 0.4807 & 0.8342 & - & - &  - & 0.4514 & 0.8038\\\hline\hline
\end{tabular}
}
\vspace{0.1cm}
\caption{Experimental results on Real-CE test set with SISR and STISR models trained on different training sets. TZ, RS and RC refer to TextZoom~\cite{wang2020scene}, RealSR~\cite{cai2019toward} and Real-CE datasets, respectively. It should be noted that the evaluated metric scores of SISR and STISR methods are very different because SISR models intake global images as input, while STISR models intake text lines as input.}
\label{table:ablation_trainingdata}
\vspace{-0.3cm}
\end{table*}




Besides the pixel-level supervision, we compute the feature level EA loss $\mathcal{L}^{F}_{EA}$ as follows:

\begin{equation}
    \mathcal{L}^{F}_{EA} = |\mathcal{F}(\hat{\mathcal{I}}_H) \cdot \mathcal{F}(\hat{\mathcal{C}}_H) - \mathcal{F}(\mathcal{I}_H) \cdot \mathcal{F}(\mathcal{C}_H)|,
\end{equation}
\noindent where $\mathcal{F}$ denotes a pretrained feature extractor network (VGG19~\cite{simonyan2014very} is used in this paper). $\mathcal{F}(\hat{\mathcal{I}}_H)$ and $\mathcal{F}({\mathcal{I}}_H)$ denote the feature representation of the estimated and the ground truth HR text images, respectively. $\mathcal{F}(\hat{\mathcal{C}}_H)$ and $\mathcal{F}({\mathcal{C}}_H)$ denote the feature representation of the estimated and the ground truth HR text edge maps, respectively. The image features are weighted by the edge features via element-wise multiplication (\eg, $\mathcal{F}(\hat{\mathcal{I}}_H) \cdot \mathcal{F}(\hat{\mathcal{C}}_H)$) to strengthen the structural areas. Finally, an $\mathcal{L}_1$ loss is imposed on the strengthened features between the estimated and ground truth ones. One can view the \textbf{supplementary file} for more detailed analysis.

Finally, together with the $\mathcal{L}_1$ loss on RGB images and EA loss terms, the overall loss function $\mathcal{L}$ is formulated as:
\begin{equation}
    \mathcal{L} =  \mathcal{L}_{1} + \alpha\mathcal{L}^{P}_{EA} + \beta\mathcal{L}^{F}_{EA},
\label{equ:allloss}
\end{equation}
where $\alpha$ and $\beta$ are balancing parameters.

\section{Experimental Results}
In this section, we first validate the effectiveness of our established Real-CE dataset by comparing STISR models trained on it and other text image datasets, and then validate the proposed EA loss in improving STISR model performance. 
All models are all trained with the Adam optimizer. When trained on our Real-CE training set, the number of epochs is set to $400$. The learning rate is set to $2 \times 10^{-4}$. In the calculation of $\mathcal{L}_{EA}$, we adopt the $Conv5\_4$ features of pre-trained VGG19~\cite{simonyan2014very}. The balancing parameters $\alpha$ and $\beta$ in Eq. (\ref{equ:allloss}) are set to $1$ and $5 \times 10^{-4}$, respectively (one can refer to \textbf{supplementary file} for the parameter selection details).
When computing recognition-based metrics, we first crop the text lines from the global text image and then rescale the cropped SR text line image to fit the recognizer.

\begin{figure*}[t]
  \centering
  \includegraphics[width=0.95\linewidth]{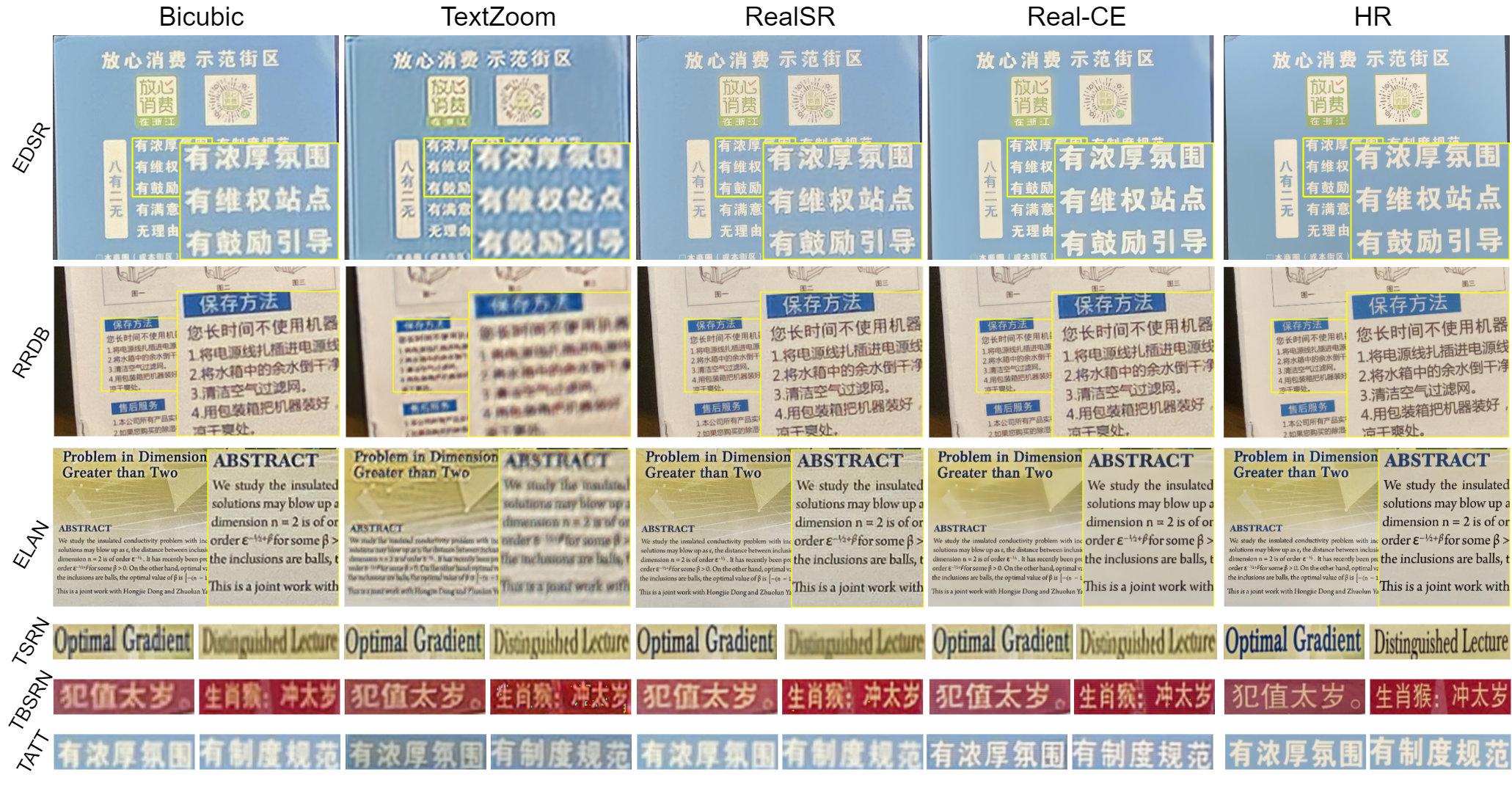}
  \caption{STISR results of different models trained on different training datasets. Note that SISR models~(EDSR~\cite{wang2018recovering}, RRDB~\cite{wang2018esrgan} and ELAN~\cite{zhang2022efficient}) intake global images as input and follow the Real-CE inference protocol, while STISR models like TSRN~\cite{wang2020scene}, TBSRN~\cite{chen2021scene} and TATT~\cite{ma2022text} can only take text line as input. }
  \label{fig:visualization}
\end{figure*}

\begin{table*}[t]
\footnotesize
\centering
\setlength{\tabcolsep}{5pt}
\scalebox{1.0}[0.95]{
\begin{tabular}{c|l|ccc|ccccc|ccccc}
\hline\hline
& SR factor & & & & \multicolumn{5}{c|}{$4\times$} & \multicolumn{5}{c}{$2\times$} \\\hline
& Approach & $\mathcal{L}_1$ & $\mathcal{L}^P_{EA}$ & $\mathcal{L}^F_{EA}$ & PSNR $\uparrow$ & SSIM $\uparrow$ & LPIPS $\downarrow$ & ACC $\uparrow$ & NED $\uparrow$ & PSNR $\uparrow$ & SSIM $\uparrow$ & LPIPS $\downarrow$ & ACC $\uparrow$ & NED $\uparrow$\\\hline
& Bicubic & - & - & - & 19.65 & 0.6684 & 0.3987 & 0.2759 & 0.6173 & 20.82 & 0.7106 & 0.2100 & 0.3475 & 0.6982\\\hline

\multirow{9}{*}{\rotatebox{90}{SISR Methods}} & \multirow{3}{*}{SRRes~\cite{ledig2017photo}}& $\checkmark$ & $\times$ & $\times$ & 20.22 & \textbf{0.7224}  & 0.2665 & 0.2879 & 0.6361 & 20.72 & 0.7360  & 0.2116 & 0.3499 & 0.6996\\
& & $\checkmark$ & $\checkmark$ & $\times$ & \textbf{20.30} & 0.7219  & 0.2722 & 0.2909 & 0.6373 & \textbf{21.23} & \textbf{0.7551}  & 0.2022 & 0.3496 & 0.7013\\
&  & $\checkmark$ & $\checkmark$ & $\checkmark$ & 20.18 & 0.7102  & \textbf{0.2041} & \textbf{0.2917} & \textbf{0.6454} & 21.09 & 0.7489  & \textbf{0.1875} & \textbf{0.3540} & \textbf{0.7080} \\\cline{2-15}
 
& \multirow{3}{*}{RRDB~\cite{wang2018esrgan}}& $\checkmark$ & $\times$ & $\times$ & 20.23 & 0.7231  & 0.2626 & 0.2920 & 0.6421 & 21.10 & 0.7535  & 0.2065 & 0.3494 & 0.7003\\
& & $\checkmark$ & $\checkmark$ & $\times$ & \textbf{20.42} & \textbf{0.7303}  & 0.2630 & 0.2914 & 0.6399 & \textbf{21.21} & \textbf{0.7559}  & 0.2010 & \textbf{0.3575} & 0.7063\\
& & $\checkmark$ & $\checkmark$ & $\checkmark$ & 20.14 & 0.7210  & \textbf{0.2031} & \textbf{0.3093} & \textbf{0.6622} & 21.00 & 0.7517 & \textbf{0.1852} & 0.3549 & \textbf{0.7130}\\\cline{2-15}
 
 
 
& \multirow{3}{*}{ELAN~\cite{zhang2022efficient}}& $\checkmark$& $\times$ &$\times$ & 20.39 & 0.7299 & 0.2892 & 0.2953 & 0.6404 & 21.16 & 0.7480 & 0.2201 & 0.3508 & 0.6992\\
& & $\checkmark$ & $\checkmark$ & $\times$ & \textbf{20.47} & \textbf{0.7330} & 0.2767 & 0.2982 & 0.6441 & \textbf{21.29} & \textbf{0.7557} & 0.1989 & \textbf{0.3549} & 0.7047\\
&  & $\checkmark$ & $\checkmark$ & $\checkmark$ & 20.21 & 0.7245 & \textbf{0.2071} & \textbf{0.3061} & \textbf{0.6567} & 21.10 & 0.7479 & \textbf{0.1835} & 0.3524 & \textbf{0.7073} \\\hline\hline

\multirow{6}{*}{\rotatebox{90}{STISR Methods}} &  \multirow{3}{*}{TBSRN~\cite{chen2021scene}} & $\checkmark$ & $\times$ & $\times$ & 18.33 & 0.4826 & 0.1715 & 0.2527 & 0.4444 & 19.01 & 0.5366 & 0.1652 & 0.3181 & 0.5294\\
& & $\checkmark$ & $\checkmark$ & $\times$ & \textbf{18.46} & \textbf{0.4881} & 0.1699 & 0.2555 & 0.4483 & \textbf{19.10} & \textbf{0.5378} & 0.1662 & 0.3185 & 0.5364\\
& & $\checkmark$ & $\checkmark$ & $\checkmark$ & 18.20 & 0.4796 & \textbf{0.1431} & \textbf{0.2615} & \textbf{0.4531}  & 18.97 & 0.5334 & \textbf{0.1307} & \textbf{0.3289} & \textbf{0.5450}\\\cline{2-15}

& \multirow{3}{*}{TATT~\cite{ma2022text}} & $\checkmark$ & $\times$ & $\times$ & 17.96 & 0.4904 & 0.1804 & 0.2330 & 0.4342 & 19.06 & 0.5772 & 0.1590 & 0.3127 & 0.5240\\
& & $\checkmark$ & $\checkmark$ & $\times$ & \textbf{18.12} & \textbf{0.4916} & 0.1786 & 0.2324 & 0.4306 & \textbf{19.17} & \textbf{0.5825} & 0.1604 & 0.3166 & 0.5360\\
& & $\checkmark$ & $\checkmark$ & $\checkmark$ & 17.89 & 0.4822 & \textbf{0.1546} & \textbf{0.2417} & \textbf{0.4549} & 19.02 & 0.5723 & \textbf{0.1475} & \textbf{0.3239} & \textbf{0.5491}\\\hline

& HR & - & - & - & - & - &  - & 0.4807 & 0.8342 & - & - &  - & 0.4514 & 0.8038\\\hline
\hline
\end{tabular}
}
\vspace{0.1cm}
\caption{Comparison of SISR and STISR models trained on Real-CE with different losses.}
\label{table:ablationRealCESR}
\end{table*}

\subsection{Effectiveness of Real-CE Dataset} 
\label{sec:exp_dataset}

In this section, we perform experiments to validate the advantages of the proposed Real-CE dataset over existing real-world SR datasets, including TextZoom~\cite{wang2020scene} and RealSR~\cite{cai2019toward}. TextZoom is built for real-world English text super-resolution, which lacks dense character structures in the dataset. RealSR is built for real-world natural image super-resolution. We evaluate five state-of-the-art SISR models and four state-of-the-art STISR models on the three datasets. The five SISR models are SRRes~\cite{ledig2017photo}, RRDB~\cite{wang2018esrgan}, EDSR~\cite{lim2017enhanced}, RCAN~\cite{zhang2018image} and ELAN~\cite{zhang2022efficient}, where the first four are CNN-based models and the last one is a transformer-based model. The four STISR models are TSRN~\cite{wang2020scene}, TPGSR~\cite{ma2021text}, TBSRN~\cite{chen2021scene} and TATT~\cite{ma2022text}, where the first two are CNN-based models, and the rest are transformer-based models. 

All the STISR and SISR models are trained on Real-CE, TextZoom and RealSR, respectively, and tested on the testing set of Real-CE. Since SISR models generally support arbitrary input sizes, the original test images are set as the input, and the PNSR, SSIM and LPIPS metrics are computed on the original sizes. Note that this is the default evaluation protocol of our benchmark, as described in Section~\ref{sec:eva}. However, most of the STISR models \cite{wang2020scene,ma2021text,chen2021scene,ma2022text} only support inputs with fixed sizes. Thus, we first crop and reshape the test images to a fixed size as the network input, and then the network outputs are compared with the resized ground truth images to compute PNSR, SSIM and LPIPS.

\begin{figure*}[t]
  \centering
  \includegraphics[width=0.9\linewidth]{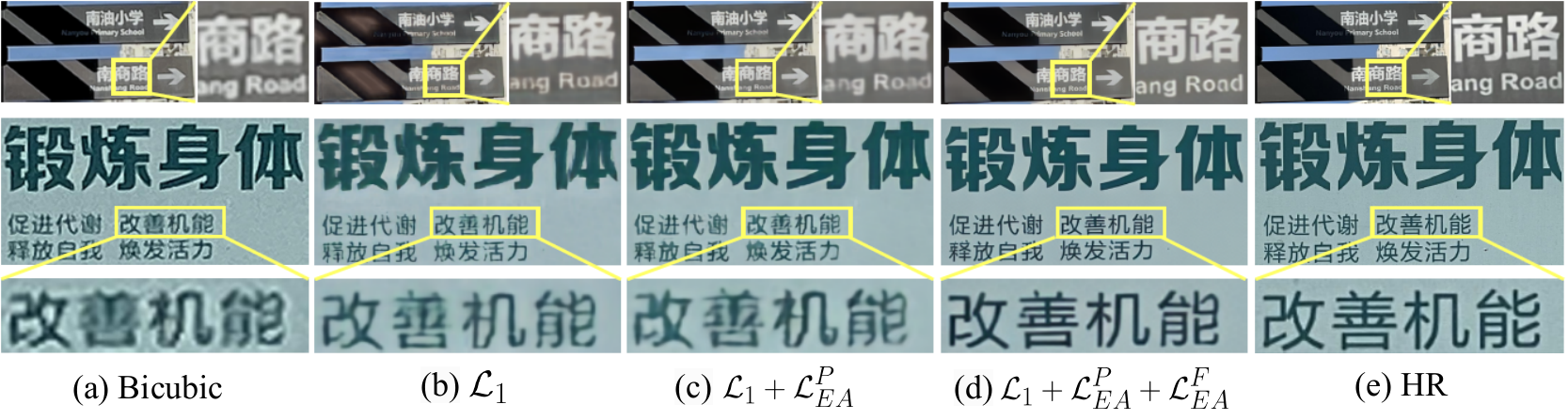}
  \caption{STISR results of RRDB models trained with different losses. }
  \label{fig:visualization_loss}
  \vspace{-0.3cm}
\end{figure*}

\begin{figure}[t]
  \centering
  \includegraphics[width=0.9\linewidth]{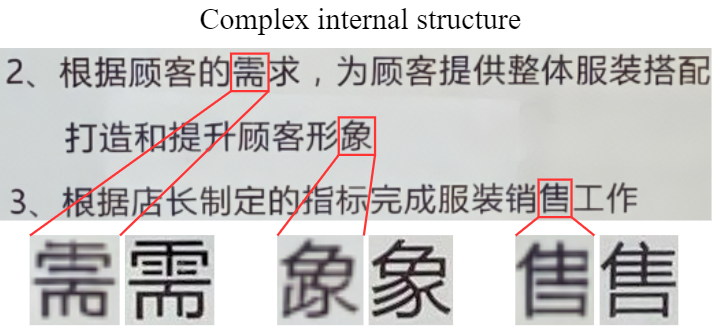}
  \caption{Examples of failure cases by our method.}
  \label{fig:failurecases}
  \vspace{-0.3cm}
\end{figure}

The quantitative results of compared SISR and STISR models are shown in Table~\ref{table:ablation_trainingdata}. 
One can see that, the models trained on TextZoom obtain inferior performance in terms of image-based metrics and recognition-based metrics. This is because the training data in TextZoom lacks complex character structures, and hence the trained models cannot handle the complex Chinese texts in Real-CE test set. Moreover, by using TextZoom, the SR models can only be trained with data of fixed sizes, which are hard to be generalized to other text image sizes. The models trained on RealSR~\cite{cai2019toward} also obtain inferior results, since RealSR is basically established for natural image SISR. In contrast, the STISR and SISR models trained on our Real-CE dataset show much better text recovery performance on all evaluation metrics. In addition, it should be noted that the evaluated metric scores of SISR and STISR methods are very different because SISR models intake global images as input, while STISR models intake text line as input. 

Figure~\ref{fig:visualization} visualizes the SR results of some representative SISR and STISR models trained on the three datasets. For convenience, we input different images to different models for more comprehensive evaluation. One can see that the text recovery results by models trained on TextZoom are blurry and contain visual artifacts. This is because TextZoom lacks training data with complex Chinese character structures. Besides, TextZoom only supports fixed-size training and the trained model cannot generalize to other sizes of testing data. The results of models trained on RealSR have fewer artifacts but are blurry with mixed strokes. In contrast, the results of models trained on Real-CE dataset have clear edges and are highly readable on both Chinese and English characters. 
More visual results can be found in the {\bf supplementary material}.

For comparisons on the STISR models trained on synthetic LR-HR data and our Real-CE data, please also refer to the {\bf supplementary material}.

\subsection{Effectiveness of the EA Loss}

We then validate the effectiveness of our proposed EA losses by testing SISR and STISR models with different combinations of $\mathcal{L}_1$, $\mathcal{L}^P_{EA}$ and $\mathcal{L}^F_{EA}$ losses. Here we employ three SISR models, including SRRes~\cite{ledig2017photo}, RRDB~\cite{wang2018esrgan} and ELAN~\cite{zhang2022efficient}, and two STISR models, including TBSRN~\cite{chen2021scene} and TATT~\cite{ma2022text}, in the experiments. The evaluation metrics are the same as that in Section~\ref{sec:exp_dataset}.

Quantitative evaluation results of the losses are shown in Table \ref{table:ablationRealCESR}. One can see that compared with the models trained with the $\mathcal{L}_1$ loss only, models trained with $\mathcal{L}_1$ and $\mathcal{L}^P_{EA}$ demonstrate enhanced PSNR/SSIM scores. This is because $\mathcal{L}^P_{EA}$ provides pixel-wise supervision on edge areas, resulting in improved pixel-wise metrics. However, the improvements on perceptual metrics (\ie, LPIPS) and recognition metrics are still limited. By further adding $\mathcal{L}^F_{EA}$ loss into training, all models demonstrate notable improvement on LPIPS and recognition accuracy, especially on $4\times$ results. This indicates that the character structural information is important for text legibility. Since the character structures can be well enhanced by using $\mathcal{L}^F_{EA}$ in training, the text recognition is significantly improved.

By using the RRDB model, we visualize the STISR results by different losses in Figure~\ref{fig:visualization_loss}. One can see that RRDB trained with only the $\mathcal{L}_1$ loss shows limited improvement compared with the bicubic interpolation. By including $\mathcal{L}^P_{EA}$ loss in training, the reconstructed text images are much enhanced with clearer character edges, as shown in Figure~\ref{fig:visualization_loss}(c). By further incorporating the $\mathcal{L}^F_{EA}$ loss, a significant enhancement in terms of edge clarity 
and local contrast can be observed, which greatly improves the legibility of Chinese text contents, as shown in Figure~\ref{fig:visualization_loss}(d). More visualization results can be found in the \textbf{supplementary file}.

\subsection{Failure Cases}

Our proposed method may fail when the character has very low resolution and intricate structures, as shown in Figure~\ref{fig:failurecases}. Though the output still has clear edges, some of the tiny strokes are wrong. This is because the tiny strokes are very obscure in the low-resolution input text image. In such cases, semantic information can be incorporated to assist the text restoration, which will be our future work. 



\section{Conclusions}

In this paper, we established a Chinese-English  benchmark, namely Real-CE, for scene text image super-resolution (STISR) model training. It contained 1,935 training and 783 testing 
images. The text region pairs contained 33,789 text lines, among which 24,666 were Chinese texts with complex structures. We further proposed an edge-aware (EA) learning method for the restoration of Chinese texts, which computed a text edge map from the given input image and employed an EA loss to guide the STISR model learning process. Experimental results demonstrated that the models trained on our Real-CE dataset can recover clearer and more readable Chinese texts than other STISR datasets, and the EA learning scheme can effectively improve text image quality. The Real-CE dataset provided a valuable benchmark for researcher to investigate the challenging Chinese text image recovery problems. 

\section{Acknowledgements}
This work is supported by the Hong Kong RGC RIF grant (R5001-18). We thank Dr. Xindong Zhang for his help on the project.


{\small
\bibliographystyle{ieee_fullname}
\bibliography{camera_ready}
}

\end{document}